\documentclass[11pt,a4paper]{article}
\usepackage{authblk}
\usepackage[hyperref]{acl2021}
\usepackage{times}
\usepackage{latexsym}

\usepackage[ruled,vlined,linesnumbered]{algorithm2e}
\aclfinalcopy % Uncomment this line for the final submission
\def\aclpaperid{***} %  Enter the acl Paper ID here
\title{Incorporating Human Explanations for Robust Hate Speech Detection}
\author[1]{\textbf{Jennifer L. Chen}}
\author[1]{\textbf{Faisal Ladhak}}
\author[1]{\textbf{Daniel Li}}
\author[2,1]{\textbf{Noémie Elhadad}}
\affil[1]{Department of Computer Science, Columbia University}
\affil[2]{Department of Biomedical Informatics, Columbia University}
\affil[ ]{\normalsize Corresponding Author: \textit {jennifer.chen@columbia.edu}}
\date{}

\begin{document}
\maketitle

\begin{abstract}
%%%%%%%%%%%%%%%%%%%%%%
% WORKSHOP LINK
% https://unimplicit.github.io/#call
%%%%%%%%%%%%%%%%%%%%%%
% it is no wonder that LM usage presents serious ethical implications
% there are many open questions about ethical concerns w/ detecting and censoring hate

%there are open questions related to the generalizability of detecting hate speech in LM-based models
% robustness or faithfulness or NLU
% detecting hate speech presents serious ethical implication

Given the black-box nature and complexity of large transformer language models (LM) \citep{bender2021dangers}, concerns about generalizability and robustness present ethical implications for domains such as hate speech (HS) detection. Using the content rich Social Bias Frames dataset \citep{biasframes}, containing human-annotated stereotypes, intent, and targeted groups, we develop a three stage analysis to evaluate if LMs faithfully assess hate speech. 
First, we observe the need for modeling contextually grounded stereotype intents to capture implicit semantic meaning.
%beyond spurious words?
Next, we design a new task, Stereotype Intent Entailment (SIE), which encourages a model to contextually understand stereotype presence.  
Finally, through ablation tests and user studies, we find a SIE objective improves content understanding, but challenges remain in modeling implicit intent.

% improves robustness to spuriousness, 
% and non-hate speech
%, demonstrating the need for careful standards in HS dataset creation and (improved modeling/training strategies for semantic coherence?)

%fall short of modeling global context for non-hate speech due to dataset spuriousness (?), demonstrating the need for careful standards in HS dataset creation

%Finally, through ablation tests and user studies on a SIE trained LM, we find a SIE objective develops holistic content understanding consistent with human reasoning and improves robustness to spuriousness. 

\end{abstract}

\section{Introduction}
Considering the pervasiveness of social media and ease of disseminating information, as well increasing divisiveness of highly contentious beliefs, political and otherwise, offensive hate speech (HS) and targeted rhetoric has unsurprisingly seen a disturbing surge in recent years \citep{mondal2017measurement, mathew2019spread, mathew2020hate}. We use the definition of HS from \citep{levy_karst_winkler_2000}\footnote{The Encyclopedia of the American Constitution} as \textit{``any communication that disparages a person or a group on the basis of some characteristic such as race, color, ethnicity, gender, sexual orientation, nationality, religion, or other characteristic"}.

It follows that combating HS through means of automatic detection and censoring has been an especially important topic of interest \citep{warner2012detecting}. While this may be well-intentioned, detecting and censoring HS poses a challenging duality: properly removing hateful content can successfully prevent violence against groups of individuals whereas over-censoring and incorrectly censoring content can place an undue burden on the freedom of speech \citep{dukes2017black, sullivan2010two, olteanu2018effect, cassidy2013cyberbullying}. 

%  ""
% TODO make this more general first, the reference curse words as example
%  For example, vulgar curse words juxtaposed with their context in hate speech classification.

% TODO mention probing to discover presence of spurious learned artifacts 
% implicit semantic roles, might be confused w/ frames

% semantic aspects ?
% TODO DO THAT EXPERIMENT
This work seeks to extend HS detection transparency and robustness. We probe an LM hate-speech detection model for artifacts indicative of spurious relations between individual words and the potentially hateful intent of an utterance. For instance, profanities, or socially offensive words, on their own are not a good proxy for HS~\citep{malmasi2018}. 
%We further assess whether this model captures aspects of hate speech (e.g., targeted groups and underlying stereotype). % TODO DO THAT EXPERIMENT
We find that a vanilla LM-based detection model places too much emphasis on artifacts. To mitigate this phenomenon, we introduce a new task, Stereotype Intent Entailment (SIE), that encourages a model to detect the semantic alignment of an utterance with respect to a given stereotype. Finally, we develop an algorithm to explore correlated word pairs and through user studies observe that the SIE-trained model learns semantic patterns related to human understanding.

\section{Hate Speech Detection and Analysis}
% Change 
% 3. Data
% INCLUDE FIGURE OF DATA. compare binary vs SIE objective
% The noisy nature of social media data warrants analysis into the robustness and learned representations of black-box LMs. Fine-tuned Roberta-base models \cite{liu2020roberta} are used for all experiments.

% We examine a transformer LM's \citep{vaswani2017attention} learned representations where we focus on contextualized embeddings (and the accompanying implicit biases that must be addressed \citep{kurita2019quantifying}) as opposed to fixed word representations\footnote{Fixed word representations still include stereotypical biases in certain words, regardless of context, due to human curated real world training data.}.
% cite word2vec, glove
% cite all of this https://www.cs.cmu.edu/~ytsvetko/papers/bias_in_bert.pdf
% To identify challenges with faithful hate speech classification, we develop three analyses based on LMs trained on our modified Social Bias Frames dataset . 
%All trained model instances are consistently trained and architecturally identical 12-layer Roberta-base models \cite{liu2020roberta} throughout all experiments.

% 

\subsection{Data and Model}
The Social Bias Frames (SBF) dataset \citep{biasframes} consists of 60k HS and lewd tweets with human-annotated labels for offensiveness, intent to offend a group, group offended, and implied stereotype sentences. For HS tweets, top 5 demographics were women (22\%), Black (21\%), assault victims (6\%), Jewish (5\%), and Muslim (4\%).

%The wealth of annotations allow for unique modifications to the dataset to study specific hypotheses. 
We consider tweets as HS if annotators labeled them as offensive and intending to offend a group (40\% HS class). \texttt{LM-HS} \footnotetext{Roberta-base finetuned with 75-25 train/test split, LR 2e-6, AdamW optimizer} achieves a test-set accuracy of 80.9\% and F1 of 84.0\%.

\subsection{Saliency Ablation Analysis}
\label{sec:bin}
To probe \texttt{LM-HS} behavior with individual word importance, we compute Input x Gradient saliency maps, which compute each token's gradient importance in a sentence to the predicted label, \citep{mudrakarta2018did} for test set sentences and evaluate single-word perturbations. 

First we perform the leave-one-out (LOO) attack \citep{feng2018pathologies}, which removes the most salient word in each test example, and find the test accuracy drops by 11\%, suggesting high concentration of predictive importance on a single word. 

Next, we test LM robustness to additional tokens that do not modify hateful intent \citet{zhong2019detecting} . We develop an adversarial adding (AA) attack by inserting top-5 highest average saliency non-negation stop words (AA-S) (e.g. `youre', `whom') and rare/misspelled words (AA-R) (e.g. `rtwoxchromosomes', `butthole',`issnowflake') of the opposite class next to the max-saliency token in each tweet. Stop and rare words in unrelated context should not modify hateful intent, but we find AA-R decreases test accuracy by almost half (38.4\%), and AA-S by 14.1\%. We also observed many HS jokes were questions, and found adding a question stop word \footnotemark \footnotetext{$\{$what, whats, how, why $\}$} (AA-Q) to the beginning of tweets dropped accuracy 19.2\% (table \ref{tab:ablation}).
 
\section{Stereotype Intent Entailment}
\label{sec:nli}
% describe data framework
Observing spuriousness in HS objective, we hypothesize that modeling contextually grounded stereotypes in a tweet encourages the model to semantically align a tweet with its underlying intent.

\subsection{Data and Model}
% Dataset statistics TODO
To adapt SBF for stereotype intent modeling, we pair HS tweets and human-annotated stereotypes as ``entailment" (stereotype is \textit{present} in the tweet). We distinguish between a stereotype being \textit{unrelated} ( ``neutral"), or \textit{opposite} (``contradict") to the underlying stereotype. ``Neutral" pairs are generated by aligning non-HS tweets with a random stereotype, and HS tweets with a stereotype of a very different demographic\footnotemark \footnotetext{stereotype group having low cosine similarity with the target group in HS tweet}. ``Contradiction" pairs are generated from HS by replacing one word in the stereotype with an antonym using Wordnet \citep{fellbaum2010wordnet} and picking high-probability antonyms with a LM scorer\footnote{GPT-2 \citep{radford2019language}}.

% User evaluation found 80\% of contradictory stereotypes retain grammatical correctness and 62\% of stereotypes are completely opposite in semantics to the original stereotype. Lower than expected agreement in stereotype-flipping was due to disagreement between users in whether changing demographic to an antonym completely flipped the meaning of the stereotype.

The final SIE dataset consists of 220k tweet \& stereotype pairs, roughly even across the three classes.  \texttt{LM-SIE} (Roberta-base) achieves F1 and accuracy of 87.6\% on the test set. 

\subsection{Tweet-Stereotype Word Pair Correlations}
% why are we developing this?
% bc it uses context less than we think, we did the NLI thing. now we ask, did the NLI thing actually align context, semantic understanding? can we visualize this?
To probe whether \texttt{LM-SIE} learns semantic alignment, we develop an algorithm that discovers globally correlated word-pairs across SIE, and evaluate pairs through users. We observe \texttt{LM-SIE} learns correlations between demographics and attributes, synonyms/antonyms, and spurious pairs.

Using Pearson correlation between train set saliency scores for each word in the tweet and each word in the stereotype, we retain word pairs with high correlation occuring in more than 3 training examples. Word pairs unique to an SIE class were also analyzed by users.

\section{Results \& Findings}
% REITERATE claims and now use actual tables and numbers
% a few quotes even maybe

We train a SOTA HS classifier and find \texttt{LM-HS} consistently overweights spurious words, surprisingly reducing accuracy by half when attacked with rare words and a quarter with question words. By incorporating human explanations of HS (stereotypes) through SIE, robustness is improved. \texttt{LM-SIE} is consistently less susceptible to attacks (table \ref{tab:ablation}), suggesting modeling explanations can improve contextual understanding.

Finally, we demonstrate SIE learns saliency-correlated word pairs that align with human understanding. Through our word-pair algorithm and user studies\footnotemark \footnotetext{10 young adults (5 female, 5 male, ages 20-28), 30 min user studies}, we find an SIE objective learns correlated tokens that helped users in reasoning about the gold SIE class 68\% of the time.
% particularly susceptible to rare/misspelled words (4x degradation of \texttt{LM-SIE}) and question stop words (7x).

% align with human understanding 
%68\% of highly correlated word pairs helped users in reasoning about the gold label SIE class. 

\begin{table}[]
\begin{tabular}{l|lllll}
          & LOO & LOO-S & AA-S & AA-R  & AA-Q \\ \hline
HS & 11.4 & 10.3  & 14.1  & 38.4  & 19.2 \\
SIE   & \textbf{5.2}  & \textbf{3.5}   & \textbf{10.6}  & \textbf{11.3} & \textbf{2.6}\\
\end{tabular}
\caption{Accuracy degradation from ablations (\ref{sec:bin}) in \texttt{LM-HS} and \texttt{LM-SE}\footnotemark. Lower is better.  \footnotetext{LOO-S \% from examples with highest saliency word as a stopword} }
\addtolength{\belowcaptionskip}{-50pt}
\label{tab:ablation}
\end{table}

\section{Conclusion \& Future Work}
In this work, we show hate speech LMs exhibit issues with robustness that can be mitigated by incorporating human explanations. Our word-pair algorithm and user studies also revealed improved content understanding by SIE. However, SIE presents limitations of requiring many annotations, and for HS in particular, lack of stereotypes for non-HS tweets limits our ability to model intents in non-HS. Given the promising initial results, we will further investigate contextualizing human explanations for robust HS classifiers through improved modeling and training. Overall, our findings demonstrate the need for careful standards in HS dataset creation and improved modeling/training strategies for semantic coherence and implicit understanding.

% User studies showed humans did not rely on individual tokens to make a HS classification 45\% of the time.

\bibliographystyle{acl_natbib}
\bibliography{acl2021}

%\appendix

\end{document}